\documentclass{llncs}
\usepackage{makeidx}  
\usepackage{amssymb}
\usepackage{mathtools}
\usepackage{graphics,graphicx}
\usepackage{algorithm}
\usepackage{algorithmic}

\begin{document}

\mainmatter              
\title{An iterative approach to Hough transform without re-voting}
\titlerunning{Piecewise HT}  
%
\author{Giorgio Ricca\inst{1} \and Mauro C. Beltrametti\inst{1} \and Anna Maria Massone\inst{2}}
\authorrunning{Giorgio Ricca et al.} 
%
\tocauthor{Giorgio Ricca, Mauro Carlo Beltrametti, Anna Maria Massone}
\institute{Dipartimento di Matematica, Universit\`a di Genova, via Dodecaneso 35,\\ I-16146 Genova, Italy\\
\and
CNR-SPIN, via Dodecaneso 33,\\ I-16146 Genova, Italy}

\maketitle              

\begin{abstract}
Many bone shapes in the human skeleton are characterized by profiles that can be associated to equations of algebraic curves. Fixing the parameters in the curve equation, by means of a classical pattern recognition procedure like the Hough transform technique, it is then possible to associate an equation to a specific bone profile. However, most skeleton districts are more accurately described by piecewise defined curves. This paper utilizes an iterative approach of the Hough transform without re-voting, to provide an efficient procedure for describing the profile of a bone in the human skeleton as a collection of different but continuously attached curves.
\end{abstract}

\section{Introduction}
The Hough transform is a well known pattern recognition technique to detect features in images when specified in some parametric form~\cite{ho62}. The classical Hough transform is most commonly used for the detection of simple curves such as straight lines, circles, ellipses~\cite{ho62,duda}. 
However, in recent papers its effectiveness has been proved in the case of more complex classes of curves, specifically, families of irreducible algebraic plane curves like elliptic curves, conchoids of Sl\"{u}se, oblique bifolium, curves with 3 convexities, Wassenaar curves, quartic curves with a tacnode, and piriform curves~\cite{BMP,etal}. Some of these curves have been successfully applied for identifying profiles of interest on CT slices of the human skeleton, including lumbar vertebrae and the spinal canal at different heights of the vertebral column~\cite{etal}. 

The basic idea to perform this identification is that points on a curve in the image space are mapped into curves (more generally hypersurfaces) in a parameter space which intersect at a uniquely determined point. The coordinates of this intersection point identify the curve in the image space. The procedure that computationally realizes this identification is based on a discretization of the parameter space and on the definition of an accumulator matrix counting for the number of points mapped into curves passing through (i.e., voting) its cells. The set of parameters corresponding to the maximum in the accumulator allows the recognition of the desired curve in the image space. 

From a mathematical viewpoint, the use of the Hough transform to recognize curves of more complexity in an image requires to prove the duality type result that associates a unique point in the parameter space to many different points on a curve in the image space~\cite{BMP}. From a computational viewpoint, the number of parameters in the curve equation defines the dimension of the parameter space. Thus, the computational effort of the voting procedure accordingly increases as the dimension of the accumulator matrix. Further, more practically, many recognition tasks in real applications require the identification of profiles that cannot be associated to a single curve but are more realistically represented by piecewise defined curves. For example, the identification of bone profiles in the human skeleton requires a method to assign an appropriate piece of curve to a specific portion of the real skeleton profile. Therefore, an approximation of the whole district shape can be obtained by using an iterative approach to Hough transform in order to iteratively approximate different parts of the same bone profile with different curves. 


In order to realize such a procedure, one can eliminate those points in the profile that have been recognized by using the Hough transform according to a specific family of curves, and re-applying the method by searching for a different curve from the family using the remaining set of points. However, this procedure is computationally expensive and therefore in this paper we introduced a {\it layered accumulator matrix} that allowed us to discard the contribution of some points in the voting procedure without re-voting.  This way, we reached the twofold objective of iteratively approximating the entire profile with a piecewise defined curve without dramatically increasing the computational cost of the algorithm.

The paper is organized as follows. In Section~\ref{HT}  we recall some theoretical results~\cite{BMP} on the Hough transform of algebraic curves. We also prove that, for the family of curves in the form used in the presented application, the requested conditions for the uniqueness of the intersection point in the parameter space are verified. In Section~\ref{piecewise} we define the layered accumulator and we present the recognition algorithm according to this new accumulator definition. In Section 4 we present some numerical results considering the recognition of a lumbar vertebra in a X-ray CT image. Finally, in Section 5 our conclusions are offered.

\section{Hough transform  of algebraic curves}
\label{HT}
This section surveys some definitions and basic facts introduced in \cite{BMP} and \cite{etal}, to which refer for more details and examples. Although these results hold over an infinite integral ring $K$, having in mind applications to imaging, we assume $K =  \bbbr$. 
We introduce a family of irreducible polynomials
\begin{equation}\label{polynomials}
f_{\lambda}(x,y) = \sum_{i,j=0}^{d} x^i y^j g_{ij}(\lambda), \;\;0 \leq i+j \leq d,
\end{equation}
where the coefficients  $g_{ij}(\lambda)$  are polynomials in the independent parameters $\lambda =(\lambda_1,\ldots,\lambda_t)$ varying in an euclidean open set $\mathcal{U} \subseteq \bbbr^t$, and with the degree, $d$, of  
$f_{\lambda}(x,y)$ not depending on $\lambda$.

Let ${\mathcal{F}}$ be the corresponding family of zero loci ${\mathcal{C}}_{\lambda}$ of $f_{\lambda}(x,y)$ and assume that each ${\mathcal{C}}_{\lambda}$ is a curve in the affine plane ${\mathbb{A}}_{(x,y)}^2(\bbbr)$. So we want a family $\mathcal{F}=\{{\mathcal{C}}_{\lambda}\}$ of irreducible curves which share the degree. If $P=(x_P,y_P)$ is a point in the image space ${\mathbb{A}}_{(x,y)}^2(\bbbr)$,
the polynomial 
\begin{equation}\label{poly-parameter}
\sum_{i,j=0}^d x^i_P y^j_P g_{ij}(\Lambda_1,\ldots, \Lambda_t) = 0, \;\; 0 \leq i+j \leq d,
\end{equation}
defines a hypersurface in the parameter space ${\mathbb{A}}_{(\Lambda_1,\ldots, \Lambda_t)}^t(\bbbr)$, where~$\Lambda_1, \ldots, \Lambda_t$ are affine indeterminates. We say that this hypersurface is the {\em Hough transform $\Gamma_P(\mathcal{F})$ of the   point $P$ with respect to the family} $\mathcal{F}$.

 In \cite[Theorem 2.2]{BMP} it is proved that, given a family ${\cal{F}}$ of irreducible curves defined as above, the Hough transforms $\Gamma_P(\mathcal{F})$, when $P$ varies on a curve ${\mathcal{C}}_{\lambda}$,  all pass through the same point $\lambda$ and, if there exists another point $\lambda'=(\lambda_1',\ldots,\lambda_t')$ where the Hough transforms $\Gamma_P(\mathcal{F})$ intersect, then ${\mathcal{C}}_{\lambda} = {\mathcal{C}}_{\lambda'}$. Moreover, the following two conditions are equivalent.
\begin{enumerate}
\item For any pair of curves ${\mathcal{C}}_{\lambda}$ and ${\mathcal{C}}_{\lambda'}$ in ${\mathcal{F}}$, the equality ${\mathcal{C}}_{\lambda} = {\mathcal{C}}_{\lambda'}$ implies $\lambda=\lambda'$.
\item For each curve ${\mathcal{C}}_{\lambda}$ in ${\mathcal{F}}$, one has $\cap_{P \in {\mathcal{C}}_\lambda} \Gamma_P(\mathcal{F})= \lambda$.
\end{enumerate}

A family ${\cal{F}}$ of curves which  meets  one of the above   equivalent conditions is said
 {\em Hough regular}. The curves of  those families  share  the same property of the straight lines as in the original definition of the Hough transform (see \cite{duda}). Therefore Hough regularity has a crucial computational role in the solution of the pattern recognition problem since for a Hough regular family of curves it is possible to formulate a computationally effective generalization of the algorithm for line detection,  able to automatically extract in the image a profile described by the curves from the family. 
 Clearly, the above equivalent properties are not true in general. For example, for the family of conics of equation $a^2 x^2 + b^2 y - 1 = 0$ we have that ${\mathcal{C}}_{a,b} = {\mathcal{C}}_{-a,-b}$. It is therefore of special interest to find Hough regular families of curves. To this aim, we observe that, if 
$$
 f_{\lambda}(x,y) = \sum_{i,j=0}^d x^i y^j g_{ij}(\lambda) \;\;\;\;
{\rm and} \;\;\;\;
 f_{\lambda'}(x,y) = \sum_{i,j=0}^d x^i y^j g_{ij}(\lambda') $$
are the equations of ${\mathcal{C}}_{\lambda}$ and ${\mathcal{C}}_{\lambda'}$ respectively, then ${\mathcal{C}}_\lambda = 
{\mathcal{C}}_{\lambda'}$ reads $f_\lambda(x,y)=kf_{\lambda'}(x,y)$ for some $k\in \bbbr^*:=\bbbr-\{0\}$, that is, 
\begin{equation}\label{CAR}
g_{ij} (\lambda) = k g_{ij}(\lambda^{\prime}),\;\;k \in \bbbr^*,
\end{equation}
for each pair of indices $i$, $j$ (see \cite[Theorem 1, p. 365]{God}). Relations (\ref{CAR}) can be used to check the Hough regularity of the family, i.e.,  that $\mathcal{C}_{\lambda}=\mathcal{C}_{\lambda'}$ implies $\lambda = \lambda'$. 

In the following example we describe the family of curves we deal with throughout the paper.

\begin{example}\label{elliptic} 
Consider the family  $\mathcal{F}$ of unbounded cubic curves of equation 
\begin{equation}\label{ell1}
 {\mathcal C}_{a,b,m,n}: y^2=m(x-n)^3-a(x-n)-b,
 \end{equation}
 for some real parameters $\lambda=(a,b, m,n)$, with $m\neq 0$. Non-singular curves from the family are elliptic curves, that is, of genus $1$. Equation (\ref{ell1}) is a 
 slight modification of the so called {\em Weierstrass form}, and allows us to better  pattern after a vertebra profile.
Let us show that the family is Hough regular. Assuming 
 ${\mathcal C}_{a,b,m,n}= {\mathcal C}_{a',b',m',n'}$,  equation (\ref{CAR}) reads
$$\begin{array}{l}
an-b-mn^3=g_{00}(\lambda)=kg_{00}(\lambda')=k(a'n'-b'-mn^3),   \\
-1=g_{02}(\lambda)=kg_{02}(\lambda')=-k,   \\
m=g_{30}(\lambda)=kg_{30}(\lambda')=km' \\
-3mn=g_{20}(\lambda)=kg_{20}(\lambda')=-3km'n'\\
3n^2m-a=g_{10}(\lambda)=kg_{10}(\lambda')=k(3n'^2m'-a').
\end{array} 
$$
  The second, third  and fourth relations clearly give $m=m'$ and $n=n'$. Then $a=a'$ by the last relation. Thus the first one yields $b=b'$.

For any point $P=(x_P,y_P)$ in the image plane ${\mathbb{A}}_{(x,y)}^2(\bbbr)$, the Hough transform of the point $P$ with respect to ${\mathcal F}$ is the quartic hypersurface in the parameter space ${\mathbb{A}}_{(A,B,M,N)}^4(\bbbr)$ of equation
\begin{equation}\label{quartica}
\Gamma_P(\mathcal{F}): M(x_P-N)^3-A(x_P-N)-B-y_P^2=0.
\end{equation}
\end{example}

\section{Piecewise Hough transform}
\label{piecewise}

Let $P_j$, $j=1,\ldots,\nu$, be the set of points of interest in the image space, and let $\Gamma_{P_j}(\mathcal{F})$ be their Hough transforms 
in the parameter space ${\mathbb{A}}_{(\Lambda_1,\ldots,\Lambda_t)}^t(\bbbr)$.
The first step of the standard recognition algorithm is based on a discretization of a subset $\mathcal{T}$ of the parameter space, i.e., on the definition of a set of cells
\begin{equation}\label{cells}
{\bf C}_{\bf n}:=\big\{\lambda \in \mathcal{T}\; \vert \;  \lambda_k\in \big[\lambda_{k,n_k}-\frac{d_k}{2}, \lambda_{k,n_k}+\frac{d_k}{2}\big), \; k=1,\ldots,t , \; n_k=1,\ldots,N_k \big\},
\end{equation}
where  $\lambda=(\lambda_1,\lambda_2,\ldots,\lambda_t)$ is a point in $\mathcal{T}$, $\lambda_{\bf n}=(\lambda_{1,n_1},\lambda_{2,n_2},\ldots, \lambda_{t,n_t})$ is a sampling point, ${\bf n}=(n_1,n_2,\ldots,n_t)$ is a multi-index identifying both a specific cell and the corresponding sampling point, $d_k$ and $N_k$ are the sampling distance and the number of samples considered along the $k$-component, respectively. 
Following the standard approach, the {\it accumulator matrix} $H=(H_{\bf n})$ is then defined as 
\begin{equation}\label{acc}
H_{\bf n}=H_{n_1,n_2,\cdots,n_t}:=\#\{P_j \;|\;  \Gamma_{P_j}(\mathcal{F})\cap {\bf C}_{\bf n}\neq\emptyset, \; 1\leq j\leq \nu\},
\end{equation}
and it is usually computed by means of the iterative process
\begin{equation}\label{accumulator-matrix}
H^{(j)}_{\bf n}= \left\{ \begin{array}{lrlr} 
H^{(j-1)}_{\bf n} +1  && {\rm if} \quad \Gamma_{P_j}({\cal{F}}) \cap {\bf C}_{\bf n} \neq \emptyset \\
H^{(j-1)}_{\bf n} && {\rm if} \quad  \Gamma_{P_j}({\cal{F}}) \cap {\bf C}_{\bf n} = \emptyset,
\end{array} 
\quad \quad j=1,\ldots,\nu
\right.
\end{equation}
with initialization $H^{(0)}_{\bf n} = 0$ and final result $H_{\bf n}=H^{(\nu)}_{\bf n}$, for each ${\bf n}$. The search for the maximum of $H$ at position ${\bf n^*}=(n_1^*,n_2^*,\ldots,n_t^*)$ allows us to identify the set of parameters $\lambda^*=\lambda_{{\bf n}^*}=(\lambda_{1,n_1^*},\lambda_{2,n_2^*},\ldots, \lambda_{t,n_t^*})$ and finally the curve ${\mathcal C}_{\lambda^*}$ best approximating the set of points $P_j$ in the image.

This approach can be effectively applied to the recognition of significant profiles in medical and astronomical images~\cite{BMP,etal}. However, 
in practical applications, a better accuracy can be achieved by approximating the profile of interest using more than one curve, each one fitting a specific portion of the profile. 

A possible approach is based on the search of successive maxima in the accumulator matrix, starting from the absolute one and moving on in descending order, up to a fixed threshold. This scheme has the main drawback that local maxima often cluster in the same region of the parameter space, thus identifying curves characterized by very similar sets of parameters, well approximating the same subsets of points in the same portion of the profile.

We propose here an alternative approach based on the principle {\it one man one vote}, i.e., each point in the dataset is allowed voting for just one curve. Once a curve has been recognized, all the points voting for it have to be removed from the dataset and the computation of the accumulator matrix has to be repeated without their contribution. In order to realize this idea without increasing the computational effort associated to the computation of the process defined by (\ref{accumulator-matrix}), we introduce a {\it layered accumulator matrix} $\mathcal{A}=(\mathcal{A}_{{\bf n},j})$, where the $j$-th layer is a binary matrix containing the votes of the $j$-th point $P_j$ of the original dataset to the whole set of cells. That is,
\begin{equation}\label{multiacc}
\mathcal{A}_{{\bf n},j}=\mathcal{A}_{n_1,n_2,\cdots,n_t,j}:=
		\begin{cases}	 
					 1, & \mbox{if } \Gamma_{P_j}(\mathcal{F})\cap {\bf C}_{\bf n}\neq\emptyset \\
					 0, & \mbox{otherwise}.
		\end{cases}
\end{equation}

By definition $\mathcal{A}_{{\bf n},j}\in \{0,1\}$  is then set to $1$ if and only if the Hough transform $\Gamma_{P_j}(\mathcal{F})$ of the point $P_j$ passes across the cell ${\bf C}_{\bf n}$. The usual accumulator $H$ can be easily computed from $\mathcal{A}$ summing over all the layers: 
\begin{equation}\label{acc_da_a}
H_{\bf n}=\sum_{j=1}^{\nu} \mathcal{A}_{{\bf n},j}, \;\;\;n_k=1,\ldots,N_k, \;\; k=1,\ldots,t.
\end{equation}
Then a first curve ${\mathcal C}_{\lambda^*}$ can be recognized in the usual way. However, the main advantage of this re-formulation is that one can easily identify the points which voted for the cell  ${\bf C}_{{\bf n}^*}$ by looking for the non-zero entries $\mathcal{A}_{{\bf n}^*,j}$,  and then remove the corresponding layers from $\mathcal{A}$. Calling $\mathcal{S}$ the whole set of values the index $j$ can assume in the original dataset, and $\mathcal{G} \subset \mathcal{S}$ the subset containing the indices of the removed layers, we can re-compute the usual accumulator $H$ without the contribution of points $P_j$, $j \in \mathcal{G}$, and without re-voting, by simply computing
\begin{equation}\label{aggH}
H_{\bf n}=\sum_{j\in \mathcal{S}\setminus \mathcal{G}}\mathcal{A}_{{\bf n},j}, \;\;\;n_k=1,\ldots,N_k, \;\; k=1,\ldots,t.
\end{equation}

\begin{figure}[!ht]
\begin{center}
\includegraphics[width=0.6\textwidth]{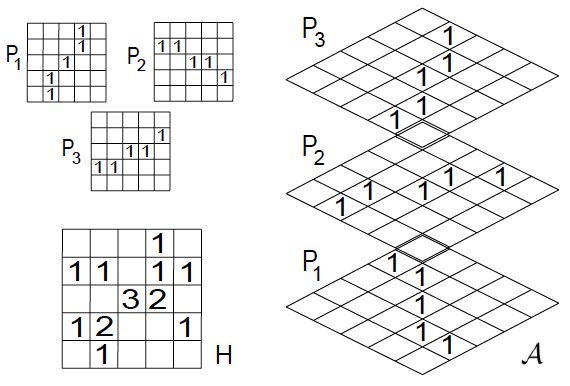}

(a) \hspace{3.2cm} (b) 
\end{center}
\caption{\small{Accumulator $H$ and layered accumulator $\mathcal{A}$ in the case of a simple dataset $P_j$, $j=1,2,3$, and a $2$-dimensional parameter space discretized in $5\times5$ cells. (a) On the top, possible votes of the three points separately. On the bottom, corresponding accumulator matrix $H$. (b) The $3$-layered accumulator matrix $\mathcal{A}$.}}
\label{fig:example_A}
\end{figure}

A new curve can now be recognized through optimization of the updated accumulator matrix, and so on. The iterative approach inspired by previous considerations is summarized in Algorithm~\ref{alg:the_alg}. A pictorial representation of the layered accumulator $\mathcal{A}$ is shown in Fig.~\ref{fig:example_A} (where we refer to the case $t=2$, so that $\mathcal{A}$ is a $3$-dimensional matrix).
\paragraph{Notes and Comments.}
Some technical details have been omitted from the previous description. For example, at each iteration we check whether the points removed from the dataset (i.e., already approximated) lie in connected components.  Moreover, as a final step, we follow an ordered path to properly connect the different branches of the piecewise defined curve. For both issues, we use arguments from graph theory~\cite{tarjan}.

\begin{algorithm}
  \caption{The Piecewise Hough transform algorithm}
  \label{alg:the_alg}
  \begin{algorithmic}[1]
  \REQUIRE A dataset of points $P_j$, $j\in \mathcal{S}$
  \ENSURE  A set of curves ${\mathcal C}_{\lambda}$ from a family ${\cal{F}}$ piecewise approximating the input dataset
    \STATE Discretization of the parameter space through a set of cells ${\bf C}_{\bf n}$ (definition~(\ref{cells}))  
    \STATE Computation of the Hough transforms $\Gamma_{P_j}(\mathcal{F})$ , $j\in \mathcal{S}$  
     \STATE Computation of the layered accumulator matrix $\mathcal{A}$ (definition~(\ref{multiacc})) 
    \STATE Computation of the accumulator matrix $H$ (equality~(\ref{acc_da_a})) 
   \WHILE {$\# \mathcal{S} > threshold\_value $}
   \STATE Search for the maximum of  $H$ at position ${\bf n^*}$
   \STATE Computation of ${\mathcal C}_{\lambda^*}={\mathcal C}_{\lambda_{{\bf n^*}}}$ 
   \STATE $\mathcal{G} \leftarrow \{j \in \mathcal{S} \; \vert \; \mathcal{A}_{{\bf n^*},j}\ne0 \}$
  \STATE Computation of $H$ from the layered accumulator $\mathcal{A}$ (equality~(\ref{aggH}))
    \STATE $\mathcal{S} \leftarrow  \mathcal{S}\setminus\mathcal{G}$
   \ENDWHILE
     \end{algorithmic}
\end{algorithm}

\section{Application to X-ray CT imaging}
\label{applicazione}

\begin{figure}[!ht]
\begin{center}
\includegraphics[width=1.\textwidth]{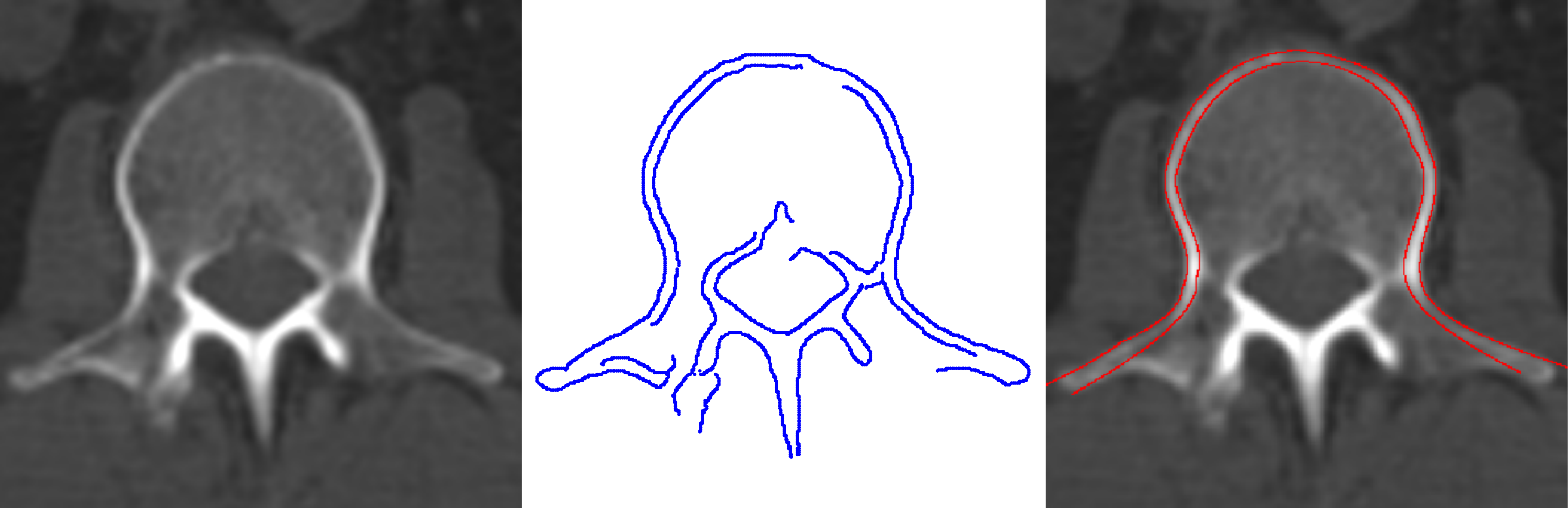}

\-  (a) \hspace{3.5cm} (b) \hspace{3.4cm} (c) 

\includegraphics[width=0.35\textwidth]{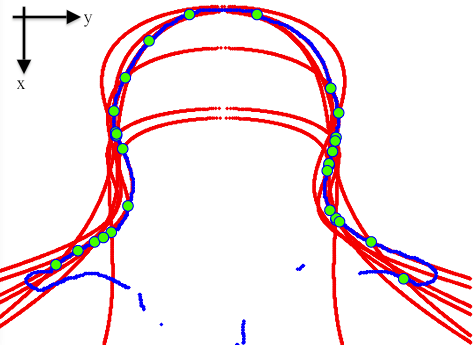}
\includegraphics[width=0.35\textwidth]{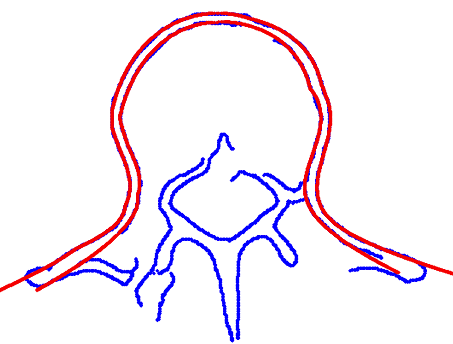}

(d) \hspace{3.8cm} (e) 
\end{center}
\caption{\small{Detection of external and internal lumbar vertebra profiles by using elliptic curves. (a) Original X-ray CT image. (b) Canny edge detection. (c) Piecewise defined curves providing the best approximation of the vertebra profiles superimposed to the original image. (d) Set of $7$ different elliptic curves piecewise approximating the edge points corresponding to the external profile. (e) Piecewise defined curves now superimposed to the whole set of edge points.}}
\label{fig:immagini}
\end{figure}

We applied the piecewise Hough transform algorithm described in the previous section to recognize a lumbar vertebra profile by using the family $\mathcal{F}$ of elliptic curves of equation (\ref{ell1}). 
In particular, we have considered an axial X-ray CT slice of a human lumbar vertebra  (see Fig.~\ref{fig:immagini}(a)) and 
we have applied a standard Canny~\cite{canny} edge detection algorithm in order to select the set of edge points shown in Fig.~\ref{fig:immagini}(b). 
From this initial dataset we then extracted the edges corresponding to the outer profile of the vertebra by applying an ad hoc {\it $\alpha$-concave hull} algorithm. This algorithm looks for the subset of points that uniquely defines a polygon enclosing the whole dataset, without having any angle between two neighboring points exceeding $\pi+\alpha$. In our application we set $\alpha=\pi/4$ and we extracted the subset $\mathcal{S}$ of $\nu=959$ blue points, as seen in Fig.~\ref{fig:immagini}(d). Each point $P_j \in \mathcal{S}$ is then Hough transformed in a quartic hypersurface of equation~(\ref{quartica}), in the parameter space ${\mathbb{A}}_{(A,B,M,N)}^4(\bbbr)$. This parameter space is then discretized in order to have a $4$-dimensional accumulator matrix $H$ made of $N_H=324135$ cells. As a consequence, the layered accumulator $\mathcal{A}$ results in a $5$-dimensional matrix with $\nu$ layers of $N_H$ entries each. The iterative scheme described in the previous section (with {\it threshold\_value} to stop the iterations set to $35\%$ of $\nu$) gives us a set of $7$ different elliptic curves from the fixed family, best approximating different portions of the outer profile of the vertebra.  The whole set of curves is superimposed to the profile edge points in Fig.~\ref{fig:immagini}(d), together with green points highlighting the boundaries of each different portion. As for the inner profile, the same process gives us a set of $9$ different curves from the family $\mathcal{F}$. The piecewise defined curves providing the best approximations for both internal and external vertebra profiles, are then shown in 
Fig.~\ref{fig:immagini}(e) and Fig.~\ref{fig:immagini}(c) on the whole set of edge points and on the original X-ray CT image, respectively. 

\section{Conclusions}
We describe an iterative approach of the Hough transform technique to provide an efficient procedure for approximating the profile of a bone in the human skeleton as a collection of different but continuously attached curves. 
The core of the proposed procedure is the definition of a layered accumulator matrix. From its knowledge, it is possible to compute the usual accumulator matrix at each iteration and to recognize and remove from the dataset the points lying on portions of the profile, already well approximated. More importantly, the layered accumulator allows us to iterate the algorithm on an input dataset progressively smaller, without the need of repeating the computationally demanding task of voting.
We tested this iterative approach to identify bone profiles in a clinical X-ray tomography image, showing that it accurately recognizes the inner and outer profiles of a human lumbar vertebra. Clinical applications of this recognition method may include either the characterization of the compact bone asset for the prognosis of chronic leukemia or the estimate of the bone marrow asset from integrated CT and nuclear medicine data~\cite{midollo}.

%

\end{document}